\input ./style/arxiv-general.cfg
\documentclass[aoas,MSNbibl,nameyear,dvips]{arximspdf}
\makeatletter
   \@ifpackageloaded{graphicx}{}{\usepackage{graphicx}}
\makeatother

\doi{10.1214/15-AOAS834}
\volume{9}
\issue{3}
\pubyear{2015}
\firstpage{1510}
\lastpage{1532}
\docsubty{FLA}

\makeatletter

\newcommand{\R}{\mathbb{R}}
\newcommand{\p}{\mathbb{P}}

\newtheorem{theorem}{Theorem}

\newproclaim{remark}[theorem]{Remark}
\makeatother

\begin{document}
\begin{frontmatter}

\title{Vertex nomination schemes for membership~prediction\thanksref{T1}}
\runtitle{Vertex nomination}
\thankstext{T1}{Supported in part by Johns Hopkins University Human Language Technology Center of Excellence (JHU HLT COE) and the XDATA program of
the Defense Advanced Research Projects Agency (DARPA) administered through Air Force Research Laboratory contract FA8750-12-2-0303.}

\begin{aug}
\author[A]{\fnms{D.~E.}~\snm{Fishkind}}, 
\author[A]{\fnms{V.}~\snm{Lyzinski}}, 
\author[A]{\fnms{H.}~\snm{Pao}}, 
\author[A]{\fnms{L.}~\snm{Chen}} 
\and
\author[A]{\fnms{C.~E.}~\snm{Priebe}\corref{}\ead[label=e5]{cep@jhu.edu}}
\runauthor{D.~E. Fishkind et al.}
\affiliation{Johns Hopkins University}
\address[A]{Department of Applied Mathematics and Statistics\\
Johns Hopkins University\\
Baltimore, Maryland 21218-2682\\
USA\\
\printead{e5}}
\end{aug}

%
\received{\smonth{8} \syear{2014}}
%
\revised{\smonth{2} \syear{2015}}

%
\begin{abstract}
Suppose that a graph is realized from a stochastic block model where
one of the blocks is of interest, but many or all of the vertices'
block labels are unobserved. The task is to order the vertices
with unobserved block labels into a ``nomination list'' such that, with high
probability, vertices from the interesting block are concentrated near the
list's beginning. We propose several vertex nomination schemes.
Our basic---but principled---setting and development yields
a best nomination scheme (which is a Bayes--Optimal analogue), and also
a likelihood maximization nomination scheme that is practical to
implement when there are a thousand vertices, and which is
empirically near-optimal when the number of vertices is small enough
to allow comparison to the best nomination
scheme. We then illustrate the robustness of the likelihood
maximization nomination scheme to the modeling challenges inherent in
real data, using
examples which include a social network involving human trafficking,
the Enron Graph, a worm brain connectome and a political blog network.
\end{abstract}

%
\begin{keyword}
\kwd{Vertex nomination}
\kwd{stochastic block model}
\kwd{graph matching}
\kwd{spectral partitioning}
\end{keyword}
\end{frontmatter}

\section{Article overview}

In a stochastic block model, the vertices of the graph are partitioned into
blocks, and the existence/nonexistence of an edge between any pair of vertices
is an independent Bernoulli trial, with the Bernoulli parameter
being a function of the block memberships of the pair
of vertices. We are concerned here with a graph realized from a
stochastic block model such that many or all of the vertices'
block labels are hidden (i.e., unobserved).
Suppose that one particular block is of interest, and the
task is to order the vertices with a hidden block label
into a ``nomination list'' with the goal of having vertices from
the interesting block
concentrated near the beginning of the list. Forming such a nomination
list can be assisted by any available knowledge about the underlying
model parameters, as well as by utilizing knowledge of block membership
for any of the vertices for which such block labels are observed.
A vertex nomination scheme is a function that, to each
such possible observed graph, assigns an associated nomination
list. In this paper we present, analyze, and illustrate the effectiveness
of several vertex nomination schemes. Some of these vertex nomination
schemes utilize graph matching and spectral partitioning machinery.
See \citet{coppersmith2014vertex}, \citet{coppersmith2012vertex} and
\citet
{lee2012bayesian} for recent work on vertex nomination, as well as a
survey of closely related problems.

One illustrative example of vertex nomination would be a social network
with vertices representing people, some of whom are engaged in human
trafficking, the rest of whom are not engaged in human trafficking, and
with edges representing a working relationship between the individuals.
Law enforcement may have as a priority separating human trafficking
from mundane sex work, because not all illegal acts represent the same
level of overall coercion. If several of these people are known to law
enforcement as human traffickers, several are known to law enforcement
to not be human traffickers, and there are very limited resources to
scrutinize the remainder as yet ambiguous people to see if they are
human traffickers, then a task would be to use the available
information and the adjacencies so as to order the as yet ambiguous
vertices into a nomination list that would prioritize these vertices
for this further scrutiny through other investigative means. In
particular, the nomination task here is a task which is not simply
classification---it is prioritization. Later, in Section~\ref{memex},
we highlight a much more elaborate real-data application of vertex
nomination in a social network involving actual human trafficking.

In Section~\ref{model} we formally and carefully define the setting
and the concept of a vertex nomination scheme.
Although prioritization is a ubiquitous need that can be treated in an
ad hoc fashion specific to individual applications,
we here formally set the problem in the stochastic block model setting,
which has gained so much popularity
in recent literature [e.g., see \citet
{NIPS20083578,Bickel15122009,nowicki2001estimation}] and is
a useful model for real data. This formal setting will be useful for
principled development of techniques that have solid theoretical
foundations and are also robust to the modeling challenges inherent in
real data.

In Section~\ref{can} we introduce the canonical vertex nomination scheme.
It is analogous to the Bayes classifier in the setting of classification.
Indeed, we prove in Proposition~\ref{optim} that the
canonical vertex nomination scheme is at least as effective as every
other vertex nomination scheme, and it thus serves the valuable role
of a ``gold standard'' with which to gauge the success of other vertex
nomination
schemes. However, it is computationally practical to implement only
when there are on the order of a very few tens of vertices.

In Section~\ref{gm} we introduce the likelihood maximization
vertex nomination scheme, which fundamentally utilizes graph matching
machinery. The graph matching problem is to find a bijection between
the vertex sets of two graphs that minimizes the number of
induced adjacency disagreements; there is a vast literature dedicated
to this problem, for example, see the article
Thirty Years of Graph Matching in Pattern Recognition [\citet{Conte2004thirty}]
for an excellent survey. Although graph matching is intractable
in theory, there have been
recent advances in approximate graph matching algorithms that are both tractable
and effective; for example, see \citet{lyzinski2013seeded}, \citet
{vogelstein2011fast} and \citet{zaslavskiy2009path}.
In particular, the very recent SGM algorithm of \citet
{lyzinski2013seeded} has
been shown in \citet{lyzinski2014graph} to be theoretically and practically
superior to convex relaxation approaches.
Using the SGM algorithm of \citet{lyzinski2013seeded} for approximate
graph matching,
the likelihood maximization vertex nomination scheme is practical to
implement for on the order of $1000$ vertices. In Sections \ref{enron},
\ref{celg} and \ref{blog}, we illustrate the robustness of
the likelihood maximization vertex nomination scheme to the model
misspecifications inherent in real data.
Furthermore, we demonstrate in Section~\ref{sims} that likelihood
maximization performs nearly as well as the canonical
``gold standard''---on graphs that have few enough vertices so
that canonical is indeed computable.

In Section~\ref{spvn} we introduce the spectral partitioning
vertex nomination scheme; it is practical to implement for
tens of thousands of vertices or more.
Based on the results in \citet{sussman2012consistent} and \citet
{fishkind2013consistent}, then followed
up in \citet{lyzinski2014perfect}, the spectral partitioning vertex nomination
scheme nominates perfectly as the number of vertices goes to infinity,
under mild conditions.

In Section~\ref{sims} we perform illustrative simulations at
three different scales, that is, a ``small scale'' experiment with
ten ambiguous vertices, a ``medium scale'' experiment with 500 ambiguous
vertices, and a ``large scale'' experiment with 10,000 ambiguous vertices.
With respect to nomination effectiveness and practicality of
implementation, the canonical vertex nomination scheme dominates
at the small scale, the likelihood maximization scheme dominates
at the medium scale, and the spectral partitioning scheme dominates
at the large scale.

In Section~\ref{enron} we illustrate
our vertex nomination schemes on the ``Enron Graph,''
a graph with email addresses of former employees of the
failed Enron Corporation as vertices, and edges indicating email contact
between the associated vertices over a time
interval. Our vertex nomination schemes are used to
nominate higher-echelon former Enron employees. Then, in Sections \ref{celg}
and \ref{blog} we illustrate on examples with a
worm-brain connectome (to nominate motor neurons) and a blog network
(to nominate political affiliation).

In Section~\ref{memex} we illustrate the impact of
our vertex nomination machinery on a real-data social network involving
human trafficking. The data are associated with
the DARPA Memex and XDATA programs.
We have a graph of web advertisements, some of them with known
association to human trafficking. Using the machinery developed in
this manuscript, we were able to nominate ambiguous advertisements
for human trafficking in a manner that was operationally significant.

\section{Vertex nomination schemes; setting and definition}
\label{model}

In this article we assume for simplicity that graphs are simple
(i.e., edges are not directed, there are no parallel edges and no
single-edge loops),
but much of what we do is generalizable.

We begin by describing the stochastic block distribution
$\operatorname{SB}(K,m,n,b,\Lambda)$,
which will be our random graph setting; its parameters are
a positive integer $K$ (the number of \textit{blocks}),
a nonnegative integer $m$ (the number of \textit{seeds}),
a positive integer $n$ (the number of \textit{ambiguous vertices}),
an arbitrary but fixed
function $b:\{ 1,2,\ldots, m+n \} \rightarrow\{1,2,\ldots,K\}$
(the \textit{block membership function}) and
a symmetric matrix $\Lambda\in[0,1]^{K \times K}$ (the \textit{adjacency
probabilities}). A random graph with distribution
$\operatorname{SB}(K,m,n,b,\Lambda)$ has the vertex set $W:=\{ 1,2,\ldots,m+n\}$
and, for each unordered pair of distinct vertices
$\{ w,w' \} \in{W \choose2}$, $w$ is adjacent to $w'$ ($w \sim w'$)
according to an independent Bernoulli trial with parameter
$\Lambda_{b(w),b(w')}$.

The vertex set $W$ is partitioned into two sets,
the set $U:=\{1,2,\ldots,m\}$ (the \textit{seeds})
and the set $V:=\{m+1,m+2,\ldots,m+n\}$
(the \textit{ambiguous vertices}).
For each $i=1,2,\ldots,K$, define $m_i:= | \{ u \in U: b(u)=i \} |$ and
$n_i:= | \{ v \in V: b(v)=i\}|$.
The function $b$ is only partially observed; its values are known
on $U$, but not on $V$. In other words, the block memberships of
the seeds are known, and the block memberships of the ambiguous vertices
are unknown, but we will assume for simplicity
that $\Lambda$ is known, and that $n_1,n_2,\ldots,n_K$ are known.
Given a random graph from $\operatorname{SB}(K,m,n,b,\Lambda)$, the most general
inferential task would be to estimate $b$ on $W$, but we will
fine tune this task very soon. (Note that if $\Lambda$ and
$n_1,n_2,\ldots,n_K$ were not known then, if there are enough seeds,
$\Lambda$ could be approximated from edge densities of
subgraphs induced by various subsets of the seeds
and, in addition, the values of $n_1,n_2,\ldots,n_K$ might be approximated
if it just so happens to be known that they are roughly
proportional to the respective values of $m_1,m_2,\ldots,m_K$. Of course,
$m_1,m_2,\ldots,m_K$ are known by virtue of the fact that $b$ is known
on $U$.)

Define $\Xi$ to be the set of bijective functions from $W$ to $W$
that fix the elements of $U$; of course, $|\Xi|=n!$. Any two graphs $G$
and $H$
on the vertex set $W$ are called \textit{equivalent} if $G$ is isomorphic
to $H$ under some function $\xi\in\Xi$; if $G$ is also asymmetric
(i.e., its automorphism group is trivial), then such a $\xi$ is
unique to $G,H$, denote it $\xi_{G,H}$.
For any graph $G$ on vertex set $W$, the equivalence class of
equivalent-to-$G$ graphs on vertex set $W$ will be denoted $\langle G
\rangle$;
in particular, $\langle G \rangle$ is an event.
The set of all such equivalence classes is denoted $\Theta$;
the events in $\Theta$ partition the sample space.

A \textit{vertex nomination scheme} $\Phi$ is a mapping that,
to each asymmetric graph $G$ with vertex set $W$, associates a linear
ordering of the vertices in $V$---called the \textit{nomination order}, and
denoted as a list $(\Phi_G(1),\Phi_G(2),\ldots,\Phi_G(n))$---such that
for every $H$ equivalent to $G$ it holds that
$(\xi_{G,H}(\Phi_G(1)),\xi_{G,H}(\Phi_G(2)),\break\ldots,  \xi_{G,H}(\Phi_G(n)))=
(\Phi_H(1),\Phi_H(2), \ldots,\Phi_H(n))$. In other words,
and descri\-bed somewhat informally, if each equivalence class of graphs
is viewed as a (single) graph whose vertex set is
comprised of labeled vertices $U$ and unlabeled vertices~$V$, then
to each equivalence class (i.e., partially vertex-labeled graph)
$\Phi$ associates a list of unlabeled vertices of $V$.

Note that the fraction of all graphs on vertex set $W$ which are symmetric
goes very quickly to zero as $|W|$ goes to infinity [\citet
{erdHos1963asymmetric,polya1937kombinatorische}].
Although
symmetric graphs are thus negligibly many, it is helpful for notation
to extend the domain of $\Phi$ to include symmetric graphs,
and this can be done in many different ways.
For simplicity of analysis we will simply say for now
that, to every symmetric graph $G$ on the vertex set $W$,
the associated nomination list is declared to be $(m+1,m+2,\ldots,m+n)$
(and we do not require the nomination list in this case to meet the
property mentioned above).

In this article, we assume that only membership
in the first block is of interest; the specific task
we are concerned with is to
find vertex nomination schemes under which there will
be, with high probability, an abundance of members of the first block
that are near the beginning of the nomination list. As an illustrative example
related to the Enron Graph example in Section~\ref{enron},
consider a corporation with $m+n=m_1+m_2+n_1+n_2$ employees,
of which $m_1+n_1$ are involved in fraud and $m_2+n_2$ are
not involved in fraud. The probability of communication between
fraudsters is fixed, as is the probability of communication
between nonfraudsters, as is the probability of communication
between any fraudster and any nonfraudster. Of the $m_1+n_1$ fraudsters,
$m_1$ have been identified as fraudsters and, among the $m_2+n_2$
nonfraudsters, $m_2$ have been identified as nonfraudsters.
Based on observing all of the employee
communications (together with knowledge of the identities of $m_1$
fraudsters and $m_2$ nonfraudsters), we wish to draw up a nomination
list of the $n_1+n_2$ ambiguous employees so that there are
many fraudsters early in the list.

The effectiveness of a vertex nomination
scheme $\Phi$ is quantified in the following manner.
For any graph $G$ with vertex set $W$,
and for any integer $j$ such that $1 \leq j \leq n$,
the \textit{precision at depth $j$} of $\Phi$ for $G$ is
defined to be $\frac{|\{ 1 \leq i \leq j :
b(\Phi_G(i))=1 \}|}{j}$;
for the corporate illustration, this represents the fraction
of the first $j$ employees on the nomination list that are
actual fraudsters in truth.
The \textit{average precision} of $\Phi$ for $G$
is defined to be $\frac{1}{n_1}\sum_{j=1}^{n_1}
\frac{|\{ 1 \leq i \leq j : b(\Phi_G(i))=1 \}|}{j}$; it
has a value between $0$ (per the corporate example,
if none of the first $n_1$ nominated employees are fraudsters)
and $1$ (if all of the first $n_1$ nominated employees are fraudsters).
Note that the average precision of $\Phi$ for $G$ is equal to
$\sum_{i=1}^{n_1} ( \frac{1}{n_1} \sum_{j=i}^{n_1} \frac{1}{j} )
\delta_{b(\Phi_G(i))=1}$, where $\delta$ is the usual indicator
function. In particular, the average
precision of $\Phi$ for $G$ is a convex combination of the indicators
$\delta_{b(\Phi_G(i))=1}$, with more weight in this convex combination
for indicators associated with lower values of $i$.
The \textit{mean average precision} of the
vertex nomination scheme $\Phi$ is the expected value of the average precision
for a random graph $G$ distributed $\operatorname{SB}(K,m,n,b,\Lambda)$.
The closer that this number is to $1$,
the more effective a vertex nomination scheme $\Phi$ is deemed.
Note that a ``chance'' vertex nomination scheme
would have the value $\frac{n_1}{n}$ as its mean average precision.

We point out that our definition of average precision is slightly
different than a definition commonly used in the
information retrieval community; our definition is a
pure average precision, whereas the other definition is actually
an integral of the precision over recall.

\section{The canonical vertex nomination scheme}
\label{can}

In this section we define the canonical vertex nomination scheme,
which is analogous to the Bayes classifier in the
Bayes classifier's setting of classification.
Indeed, we prove in Proposition~\ref{optim} that the mean average
precision of the canonical vertex nomination scheme is
greater than or equal to the mean average precision
of every other vertex nomination scheme. Unfortunately, because of its
computational intractability (a visibly exponential runtime as
the number of vertices increases), the canonical vertex nomination scheme
is only practical to implement for up to a few tens of vertices.
Nonetheless, because of Proposition~\ref{optim}, the canonical
vertex nomination scheme serves as a valuable ``gold standard'' to
evaluate the
performance of other more computationally tractable vertex nomination schemes.
(This is analogous to the role of the Bayes classifier in the
classification setting.) Our ongoing research seeks to approximate the
canonical vertex nomination scheme in a scalable fashion.

\subsection{Definition of the scheme}

Consider the random graph $G$ distributed
$\operatorname{SB}(K,m,n,b,\Lambda)$. When $G$ is asymmetric then, for any $v \in V$,
the conditional probability
%
\begin{equation}
\label{cpb} \p\bigl[ \bigl\{ H \in\langle G \rangle: b\bigl(
\xi_{G,H}(v) \bigr)=1 \bigr\} | \langle G \rangle\bigr]
\end{equation}
may be described as the probability, given
the event that we observe a graph equivalent to $G$,
that the vertex corresponding to $v$ would be in the first block.
The \textit{canonical vertex nomination scheme},
which we denote as $\Phi^C$, orders the vertices of $V$ as
$\Phi^C_G(1), \Phi^C_G(2), \ldots, \Phi^C_G(n)$
in decreasing order of this conditional probability; that is, we
define $\Phi^C$ so that,
for all $i=1,2,\ldots,n-1$,
%
\begin{eqnarray}
\label{othr}&& \p\bigl[ \bigl\{ H \in\langle G \rangle: b\bigl(
\xi_{G,H}\bigl( \Phi_G^C(i) \bigr) \bigr)=1
\bigr\} | \langle G \rangle\bigr]
\nonumber
\\[-8pt]
\\[-8pt]
\nonumber
&&\qquad\geq\p\bigl[ \bigl\{ H \in\langle G \rangle: b\bigl( \xi_{G,H}\bigl(
\Phi_G^C(i+1) \bigr) \bigr)=1 \bigr\} | \langle G \rangle
\bigr].
\end{eqnarray}

To more easily compute the conditional probability
in equation (\ref{cpb}), let ${V \choose n_1,n_2,\ldots,n_K }$
denote the collection of all the ${n \choose n_1,n_2,\ldots,n_k}$
partitions\vspace*{1.5pt} of the
elements of $V$ into subsets called $V_1, V_2, \ldots,
V_K$ with respective cardinalities $n_1,n_2,\ldots,n_K$.
Given any such partition $(V_1, V_2, \ldots,V_K) \in
{V \choose n_1,n_2,\ldots,n_K }$, let us create the following notation.
For any $k=1,2,\ldots,K$ and $\ell=k+1,k+2,\ldots,K$,
let $e_{k,\ell}$ denote the number of edges in $G$ with one
endpoint in $V_k \cup\{ u \in U: b(u)=k \}$ and the other endpoint
in $V_\ell\cup\{ u \in U:b(u)=\ell\} $, and
define $c_{k,\ell}:=(m_k+n_k)(m_\ell+n_\ell)-e_{k,\ell}$.
Let $e_{k,k}$ denote the number of edges in $G$ with both endpoints in
$V_k \cup\{ u \in U: b(u)=k \}$, and define
$c_{k,k}:={m_k+n_k \choose2}-e_{k,k}$.
Then, in the stochastic block model,
the conditional probability in equation (\ref{cpb}) can be
computed as
%
\begin{eqnarray}
& & \frac{
\sum_{(V_1, V_2, \ldots, V_K)\label{condprob}
\in{V \choose n_1,n_2,\ldots,n_K}\
\mathrm{such\ that\ }v \in V_1}
\prod_{k=1}^K \prod_{\ell=k}^K
(\Lambda_{k,\ell})^{e_{k,\ell}}
(1-\Lambda_{k,\ell})^{c_{k,\ell}}}{
\sum_{
(V_1, V_2, \ldots, V_K)
\in{V \choose n_1,n_2,\ldots,n_K}}
\prod_{k=1}^K \prod_{\ell=k}^K
(\Lambda_{k,\ell})^{e_{k,\ell}}
(1-\Lambda_{k,\ell})^{c_{k,\ell}}}.\hspace*{-20pt} 
\end{eqnarray}
Although we are not able to evaluate the probability of
$G$ since the block membership function $b$ is not fully observed,
nonetheless, the conditional probabilities in equation (\ref{cpb}) can
indeed be evaluated via equation (\ref{condprob}) by just knowing the
values of the parameters $n_1,n_2, \ldots, n_K$ and $\Lambda$.

\subsection{Optimality of the canonical vertex nomination scheme}

\begin{theorem} \label{optim}
For any vertex nomination scheme $\Phi$, the mean average precision
of the canonical vertex nomination scheme
$\Phi^C$ is greater then or equal to the mean average precision of
$\Phi$.
\end{theorem}

\begin{pf}
For each $i=1,2,\ldots,n_1$, define
$\alpha_i:=\frac{1}{n_1} \sum_{j=i}^{n_1}\frac{1}{j}$ and,
for each $i=n_1+1,n_1+2,\ldots,n$, define $\alpha_i:=0$.
The sequence $\alpha_1,\alpha_2,\ldots,\alpha_n$ is clearly
a nonnegative, nonincreasing sequence.
Note that if $a_1,a_2,\ldots,a_n$ is any
(other) nonincreasing, nonnegative sequence of real numbers, and
$a'_1,a'_2,\ldots,a'_n$ is any permutation of the sequence
$a_1,a_2,\ldots,a_n$, then
%
\begin{equation}
\label{use} \sum_{i=1}^n
\alpha_i a'_i \leq\sum
_{i=1}^n\alpha_i a_i.
\end{equation}
Indeed, this is easily verified by first considering
particular sequences $a_1,a_2,\break\ldots, a_n$ of the form
$1,1,\ldots,1,0,\ldots,0,0$ (i.e., $j$ consecutive $1$'s followed by
$n-j$ consecutive $0$'s, for different values of $j=1,2,\ldots,n$)
and then noting that the nonnegative combinations of such particular
sequences indeed comprise all nonincreasing, nonnegative sequences with
$n$ entries.

Consider the random graph $G$ distributed
$\operatorname{SB}(K,m,n,b,\Lambda)$. Recall that $\Theta$ denotes the
set of equivalence classes of graphs on the vertex set $W$.

Expanding the mean average precisions of $\Phi$, then bounding
and simplifying, yields
%
\begin{eqnarray}\label{one}
\nonumber
\mathbb{E} \Biggl( \sum_{i=1}^n
\alpha_i \delta_{ b(\Phi_G(i))=1} \Biggr) &=& \sum
_{i=1}^n \alpha_i \mathbb{P} \bigl(b
\bigl(\Phi_G(i)\bigr)=1 \bigr)
\\
\nonumber
& = & \sum_{i=1}^n
\alpha_i \biggl( \sum_{\mathcal{G} \in\Theta} \mathbb{P} (
\mathcal{G} ) \mathbb{P} \bigl( b\bigl(\Phi_G(i)\bigr)=1 \Big|
\mathcal{G} \bigr) \biggr)
\\
& = & \sum_{ \mathcal{G} \in\Theta} \mathbb{P} ( \mathcal{G}
) \Biggl( \sum_{i=1}^n
\alpha_i \mathbb{P} \bigl( b\bigl(\Phi_G(i)\bigr)=1 \Big|
\mathcal{G} \bigr) \Biggr)
\\
\nonumber
& \leq&  \sum_{ \mathcal{G} \in\Theta} \mathbb{P} (
\mathcal{G} ) \Biggl( \sum_{i=1}^n
\alpha_i \mathbb{P} \bigl( b\bigl(\Phi^C_G(i)
\bigr)=1 \Big| \mathcal{G} \bigr) \Biggr)
\\
\nonumber
& = & \sum_{i=1}^n
\alpha_i \mathbb{P} \bigl( b\bigl(\Phi^C_G(i)
\bigr)=1 \bigr) = \mathbb{E} \Biggl( \sum_{i=1}^n
\alpha_i \delta_{ b(\Phi^C_G(i))=1 } \Biggr),
\end{eqnarray}
where the inequality in equation (\ref{one}) follows from
equations (\ref{use}) and (\ref{othr}),
(and from our assumption that all nomination schemes
agree when $G$ is symmetric).
The desired result is shown.
\end{pf}

\section{Likelihood maximization vertex nomination scheme}
\label{gm}

In this section we define the likelihood maximization
vertex nomination scheme. It will be practical to implement
even when there are on the order of a thousand vertices. We will see
in Section~\ref{sims} that it is a very effective
vertex nomination scheme, when compared to the canonical
vertex nomination scheme ``gold standard'' on graphs small enough to make
the comparison. In Sections \ref{enron},
\ref{celg} and \ref{blog} we will see that likelihood maximization
appears to be nicely robust to the modeling challenges inherent in
real data.

\subsection{Definition of the scheme}
\label{schmdef}

Suppose the random graph $G$ is distributed
$\operatorname{SB}(K,m,n,b,\Lambda)$.
There are two stages in defining---and computing---the
likelihood maximization vertex nomination scheme.

The first stage is concerned with estimating the block assignment
function $b$. Let $\mathfrak{B}$ denote the set of functions
$\mathfrak{b} : W \rightarrow\{ 1,2,\ldots,K \}$ such that
$\mathfrak{b}$ agrees with $b$ on $U$, and such that it also holds,
for all $i=1,2,\ldots,K$,
that $| \{ v \in V: \mathfrak{b} (v)=i\} |=n_i$.
For any $\mathfrak{b} \in\mathfrak{B}$, and
for all $k=1,2,\ldots,K$ and $\ell=k+1,k+2$, $\ldots,K$, let
$e_{k,\ell}(\mathfrak{b})$ denote the number of edges in $G$
with one endpoint in $\{ w \in W : \mathfrak{b}(w)=k \}$ and the
other endpoint in $\{ w \in W : \mathfrak{b}(w)=\ell\}$, and also
denote $c_{k,\ell}(\mathfrak{b}):=(m_k+n_k)(m_\ell+n_\ell
)-e_{k,\ell
}(\mathfrak{b})$.
For all $k=1,2,\ldots,K$,
let $e_{k,k}(\mathfrak{b})$ denote the number of edges in $G$ with both
endpoints in $\{ w \in W: \mathfrak{b}(w)=k \}$, and also denote
$c_{k,k}(\mathfrak{b}):={m_k+n_k \choose2}-e_{k,k}(\mathfrak{b})$.
In the $\operatorname{SB}(K,m,n,b,\Lambda)$ distribution, if $b$ had been replaced with
$\mathfrak{b} \in\mathfrak{B}$, then the probability of
realizing the graph $G$ would have been
%
\begin{equation}
p(\mathfrak{b},G):=\prod_{k=1}^K \prod
_{\ell=k}^K (\Lambda_{k,\ell
})^{e_{k,\ell}(\mathfrak{b})}
(1-\Lambda_{k,\ell})^{c_{k,\ell}(\mathfrak{b})}.
\end{equation}
Define $\hat{b}$,
the \textit{maximum likelihood estimator of} $b$,
to be the member of $\mathfrak{B}$ such that the probability
of $G$ is maximized. In other words (then taking logarithms
and ignoring additive constants),
%
\begin{eqnarray}
\label{nex}
\hat{b} & :=& \arg\max_{ \mathfrak{b} \in\mathfrak{B}} p(
\mathfrak{b},G) =  \arg\max_{\mathfrak{b}\in\mathfrak{B}} \sum
_{k=1}^K \sum_{\ell=k}^K
e_{k,\ell}(\mathfrak{b}) \log\biggl( \frac{\Lambda_{k,\ell
}}{1-\Lambda_{k,\ell}} \biggr)
\nonumber
\\[-8pt]
\\[-8pt]
\nonumber
& = & \arg\max_{\mathfrak{b}\in\mathfrak{B}} \sum_{ \{w,w'\} \in{W
\choose2} }
\delta_{w \sim_G w'} \log\biggl( \frac{\Lambda_{\mathfrak
{b}(w),\mathfrak{b}(w')}}{1-
\Lambda_{\mathfrak{b}(w),\mathfrak{b}(w')}} \biggr).
\end{eqnarray}
The optimization problem in equation (\ref{nex})
is an example of seeded graph matching, and
we can efficiently and effectively approximate its solution.
The details of this are deferred to the next section, Section~\ref{sgmsol},
and we now continue on to the second stage of defining and computing the
likelihood maximization vertex nomination scheme, assuming
that we have computed~$\hat{b}$.

For any $v,v' \in V$ such that $\hat{b}(v)=1$ and $\hat{b}(v') \ne1$,
define $\hat{b}_{v \leftrightarrow v'} \in\mathfrak{B}$ such that
$\hat{b}_{v \leftrightarrow v'}$ agrees with $\hat{b}$ for all $w \in W$
except that $\hat{b}_{v \leftrightarrow v'}(v')=1$ and
$\hat{b}_{v \leftrightarrow v'}(v)=\hat{b}(v')$.
For any $v,v' \in V$ such that $\hat{b}(v)=1$ and $\hat{b}(v') \ne1$,
we can interpret a low/high value of
the quantity $\frac{p(\hat{b}_{v \leftrightarrow v'},G)}{p(\hat{b},G)}$
as a measure of our conviction/lack-of-conviction that $\hat{b}$ should
be used to estimate $b$, as opposed to estimating $b$ with
specifically $\hat{b}_{v \leftrightarrow v'}$.
In this spirit, for all $v \in V$ such that $\hat{b}(v)=1$, a low/high
value of the geometric mean
%
\begin{equation}
\label{thu} \biggl( \prod_{v' \in V: \hat{b}(v')\ne1} \frac
{p(\hat{b}_{v \leftrightarrow v'},G)}{p(\hat{b},G)}
\biggr)^{{1}/{(n-n_1)}}
\end{equation}
can be interpreted as a measure (for the purpose of ordering)
of our conviction/\break lack-of-conviction in our estimation that $b(v)$ is $1$.
Also, for all $v' \in V$ such that $\hat{b}(v') \ne1$, a low/high
value of the geometric mean
%
\begin{equation}
\label{thv} \biggl( \prod_{v \in V: \hat{b}(v)= 1} \frac{p(\hat
{b}_{v \leftrightarrow v'},G)}{p(\hat{b},G)}
\biggr)^{{1}/{n_1}}
\end{equation}
can be interpreted as a measure (just for the purpose
of ordering) of our conviction/lack-of-conviction
in our estimation that $b(v')$ is not $1$.

We now define the
\textit{likelihood maximization
vertex nomination scheme} $\Phi^L$ to be such that it satisfies
$\Phi^L_G(1),\Phi^L_G(2),\ldots,\Phi^L_G(n_1)$ are the $v \in V$
such that $\hat{b}(v)=1$, listed in increasing order of the geometric mean
in equation (\ref{thu}), and
$\Phi^L_G(n_1+1),\Phi^L_G(n_1+2),\ldots,\Phi^L_G(n)$ are the $v'
\in V$
such that $\hat{b}(v') \ne1$, listed in decreasing order of the
geometric mean
in equation (\ref{thv}).

\subsection{Solving the seeded graph matching problem}
\label{sgmsol}

In this section we discuss how to compute $\hat{b}$ in the likelihood
maximization
vertex nomination scheme $\Phi^L$ defined in the previous section.

Given any $A,B \in\R^{(m+n)\times(m+n)}$, the \textit{quadratic assignment
problem} is to minimize $\|A-PBP^T\|_F^2$ over all permutation matrices
$P \in\{ 0,1 \}^{(m+n)\times(m+n)}$, where $\| \cdot\|_F$ denotes the
Frobenius matrix norm. If $A$ and $B$ are, respectively, adjacency matrices
for two graphs, then this is called the \textit{graph matching problem};
it is clearly equivalent to finding a bijection from the vertex set of
one graph to the vertex set of the other graph so as to minimize the
number of adjacency disagreements induced by the bijection.
If $P$ is further constrained so that the upper left corner
is the $m \times m$ identity matrix, then the problem is called the
\textit{seeded quadratic assignment problem}/\textit{seeded graph matching
problem}; for graphs, this further restriction just means that
part of the bijection between the vertex sets is fixed.

Note that the objective function can be simplified (under the
restriction that $P$ is a permutation matrix) as
$\| A -PBP^T\|_F^2 = \|A \|_F^2 + \|B \|_F^2 -2 \langle A, PBP^T
\rangle$,
where $\langle\cdot, \cdot\rangle$ is the usual inner
product $\langle C,D \rangle:=\sum_{i,j}C_{ij}D_{ij}$. Thus, the
above problems
can be phrased as maximize $\langle A, PBP^T \rangle$ over all permutation
matrices $P$.

The optimization problem in equation (\ref{nex}), for which
$\hat{b}$ is the solution, is precisely the seeded quadratic assignment
problem above, where $A \in\R^{(m+n)\times(m+n)}$ is the
adjacency matrix for the graph $G$, that is, $A_{i,j}:=\delta_{i \sim
_G j}$
for all $i,j \in W \equiv\{ 1,2,\ldots,m+n \}$, and
$B \in\R^{(m+n )\times(m+n)}$ is the matrix wherein
$B_{i,j}:=\log(
\frac{ \Lambda_{\mathfrak{b}'(i),\mathfrak{b}'(j)} }{1- \Lambda
_{\mathfrak{b}'(i),\mathfrak{b}'(j)} } )$
for all $i,j \in W$, where $\mathfrak{b}'$ is the member of
$\mathfrak{B}$ for which the sequence
$\mathfrak{b}'(m+1),\mathfrak{b}'(m+2),\ldots, \mathfrak{b}'(m+n)$
are $1$'s contiguously, then $2$'s contiguously, \ldots, then $K$'s
contiguously.
The $\mathfrak{b} \in\mathfrak{B}$---over
which the objective function in equation (\ref{nex})
is maximized---correspond precisely
to the permutation matrices $P$ in the seeded quadratic
assignment problem, where the upper left corner of $P$ is restricted
to be the $m \times m$ identity matrix. We will call this problem
a seeded graph matching problem because $A$ is an adjacency matrix.
(And we can also choose to think of $B$ as a weighted adjacency matrix
for a graph.)

The seeded graph matching problem is computationally hard; indeed,
the quadratic assignment problem is NP-hard, and even deciding
if two graphs are isomorphic is notoriously of unknown
complexity [\citet{garey1979computer}, \citet{read1977graph}].
However, approximate solutions can be found efficiently with the SGM
(Seeded Graph Matching) Algorithm of \citet{lyzinski2013seeded}, which
is a
seeded version of the FAQ algorithm of \citet{vogelstein2011fast}.
[Indeed, SGM is more effective than convex relaxation techniques,
as was recently shown in \citet{lyzinski2014graph}.]
We employ the SGM algorithm to obtain an approximate
solution to $\hat{b}$ for use in the likelihood maximization
vertex nomination scheme. It runs in time $O(n^3)$, and can
be implemented even when $n$ is approximately $1000$.

\section{The spectral partitioning vertex nomination scheme}
\label{spvn}

In this section we introduce the \textit{spectral partitioning
vertex nomination scheme}. Suppose $G$ is distributed
$\operatorname{SB}(K,m,n,b,\Lambda)$.
We do not need to assume here that we know $n_1,n_2,\ldots,n_K$,
nor the entries of $\Lambda$;
we just need to know the value of $K$ and $d:=$ the rank of $\Lambda$.
[Indeed, by the results in \citet{fishkind2013consistent}, even just knowing
an upper bound on $d$ will be sufficient
to obtain good performance.]

Say that the adjacency matrix for $G$ is
$A \in\{ 0,1 \}^{(m+n) \times(m+n)}$,
that is, $A_{i,j}:=\delta_{i \sim_G j}$
for all $i,j \in W \equiv\{ 1,2,\ldots,m+n \}$.
Compute $d$ eigenvectors associated, respectively,
with the $d$ largest-modulus eigenvalues of $A$.
Scale these eigenvectors so that their respective lengths are the
square roots of the absolute values of their corresponding eigenvalues,
and define $X \in\mathbb{R}^{(m+n) \times d}$ to have these
scaled eigenvectors as its respective columns. The rows of $X$ are
low-dimensional
embeddings of the corresponding vertices.
Now, cluster the rows of $X$ into $K$ clusters; that is, solve
the problem minimize $\| X - C \|_F$ over all
matrices $C \in\mathbb{R}^{(m+n) \times d}$
with the property that each row of $C$ is equal to one of just $K$
row vectors, and the values of these $K$ row vectors
are also variables to be optimized over.

Say that $c$ is the most frequent value of row vector
in the optimal $C$ among the rows corresponding to the vertices $\{
u \in U: b(u)=1\}$.
(In other words, $c$ is the centroid associated with the most
vertices known to be in the first block.)
The spectral partitioning vertex nominating scheme, denoted by
$\Phi^S$, associates with $G$ the
ordering (of vertices in $V$)
$\Phi^S_G(1), \Phi^S_G(2),\ldots,
\Phi^S_G(n)$
in increasing order of Euclidean distance between $c$
and their corresponding row in $X$.

Suppose we consider
a sequence of graphs realized from
the distributions $\operatorname{SB}(K,m,n,b,\Lambda)$
for, successively,
$m+n=1,2,3,\ldots,$ where $K$ and $\Lambda$
are fixed, and $\Lambda$ is positive semi-definite
with the property that no two of its
rows are equal. Also, assume that
$m_1 \geq1$, and there exists a positive constant $\gamma$ such that,
for all $i=1,2,\ldots,K$, it holds that $m_i+n_i \geq\gamma
(m+n)^{{3}/{4}+\gamma}$.
It was recently shown in \citet{lyzinski2014perfect} [following the
work in \citet{sussman2012consistent} and \citet{fishkind2013consistent}]
that almost surely there
are no incorrectly clustered vertices in the limit. This
implies that the mean average precision of $\Phi^S$
converges to $1$ as $m+n \rightarrow\infty$.

It will be computationally convenient to approximately
(but very quickly) solve the clustering subproblem.
This approximate clustering can be done with the
$k$-means algorithm or with the mclust procedure [\citet
{fraley1999mclust,fraley2003enhanced}].
In both cases, the vertices are nominated
based on distance to cluster centroids; in
$k$-means this amounts to the usual Euclidean distance, while
for mclust this amounts to nominating based on the Mahalonobis distance.

\section{The OTS vertex nomination scheme}

The chief contribution of this manuscript is the formulation
of the likelihood maximization vertex nomination scheme, along
with our demonstration of its effectiveness; indeed, it is comparably
effective to the ``gold standard'' canonical vertex nomination scheme
(on graphs small enough to practically make this comparison,
as we demonstrate in Section~\ref{sims}) and it is
relatively robust to pathologies inherent in real data (as
we demonstrate later in Section~\ref{realdata}).

However, it is worthwhile to point out that classification
algorithms for stochastic block models can often be naturally
modified for use in nomination, by utilizing algorithm-inherent
numeric scores to perform vertex ranking.
For an excellent survey of the literature
on community detection in networks---including the setting of
stochastic block models---and available algorithms,
see the very comprehensive survey article \citet
{fortunato2010community} and
the papers cited therein, such as \citet{newman2004finding}
and the classic article \citet{nowicki2001estimation}.
Also, see Latent Dirichlet Allocation (LDA) [\citet
{Blei:2003:LDA:944919.944937}].
Because of the vast number of citations to it in the literature, we
next choose to
focus on the paper \citet{NIPS20083578}, titled ``Mixed membership stochastic
blockmodels,'' and the associated
R code which we call ``MMSB'' located at
\surl{http://cran.r-project.org/web/packages/lda/lda.pdf} [\citet{chang1lda}];
in the setting of a mixed membership block model,
MMSB assigns to each vertex a posterior probability of block
membership in each of the various blocks. With this, we now define
the \textit{OTS vertex nomination scheme}, denoted $\Phi^O$, which
uses MMSB to order the vertices of $V$ in decreasing order of
posterior probability of membership in the specific block indicated
by the most seeds.

We call this nomination scheme OTS ``Off The Shelf'' to emphasize that
we use MMSB as a black box without getting under the hood of the code;
as such, the use of the seeds is only to identify the
block of interest. Indeed, under the hood modifications of existing
community detection algorithms such as
MMSB and LDA and LDA-based methodologies are expected to yield new
vertex nomination schemes that will be increasingly effective and fast.
We also expect even more effective vertex nomination schemes to come
from merging vertex nomination techniques, perhaps similar
in spirit to the work in \citet{lyzinski2013spectral}, where graph matching
and spectral partitioning are merged into a more effective
avenue of graph matching for large graphs.

\section{Simulations: Comparing the vertex nomination schemes at three
different
scales}
\label{sims}

In this section we compare and contrast these vertex
nomination schemes using three simulation experiments---essentially
the same experiment at three different scales, ``small scale,''
``medium scale''
and ``large scale.''
For each of the three experiments,
we have $K=3$ blocks in the stochastic block model.
The matrix of Bernoulli parameters $\Lambda$ is
\[
\Lambda(\vartheta) := \vartheta\left[ \matrix{ 0.5 & 0.3 &
0.4
\vspace*{2pt}\cr
0.3 & 0.8 & 0.6
\vspace*{2pt}\cr
0.4 & 0.6 & 0.3}
\right] +(1-\vartheta) \left[
\matrix{ 0.5 & 0.5 & 0.5
\vspace*{2pt}\cr
0.5 & 0.5 & 0.5
\vspace*{2pt}\cr
0.5 & 0.5 & 0.5
}
\right],
\]
with the value $\vartheta=1$ for the small scale experiment,
$\vartheta=0.3$ for the medium scale experiment, and
$\vartheta=0.1$ for the large scale experiment, in order
to decrease the signal when the number of vertices is larger.

Specifically, the matrix $\Lambda$ for the small scale experiment, for the
medium scale experiment and for the large scale experiment are, respectively,
\begin{eqnarray*}
\Lambda(1)&=& \left[ \matrix{ 0.5 & 0.3 & 0.4
\vspace*{2pt}\cr
0.3 & 0.8 & 0.6
\vspace*{2pt}\cr
0.4 & 0.6 & 0.3}
\right], \qquad \Lambda(0.3)= \left[
\matrix{ 0.50 & 0.44 & 0.47
\vspace*{2pt}\cr
0.44 & 0.59 & 0.53
\vspace*{2pt}\cr
0.47 & 0.53 & 0.44}
\right],\\
\Lambda(0.1) &=& \left[
\matrix{ 0.50 & 0.48 & 0.49
\vspace*{2pt}\cr
0.48 & 0.53 & 0.51
\vspace*{2pt}\cr
0.49 & 0.51 & 0.48}
\right],
\end{eqnarray*}
so that as the number of vertices increases we have that
$\vartheta$ gets closer to zero, which means that the blocks become
less and
less stochastically differentiable one from the other. Another
notable feature of the $\Lambda$ here is that the block of
interest---the first block---is the intermediate density block, that
is, the Bernoulli adjacency parameter for vertices in the first
block is between the Bernoulli adjacency parameter for vertices in the
second block and in the third block. This makes it more challenging to
identify the
vertices of the first block, which is the
block of interest.

The values of $(n_1,n_2,n_3)$ are taken to be multiples of $(4,3,3)$,
specifically, in the small-scale experiment $(n_1,n_2,n_3)=(4,3,3)$,
in the medium-scale experiment $(n_1,n_2,n_3)=(200,150,150)$ and in the
large-scale experiment $(n_1,n_2,n_3)=(4000,3000,3000)$.
As for the seeds, the values of $(m_1,m_2,m_3)$ in the respective
experiments were taken as $(4,0,0)$, $(20,0,0)$ and\break $(40,0,0)$.

%
%

\begin{figure}
\centering
\begin{tabular}{@{}c@{}}

\includegraphics{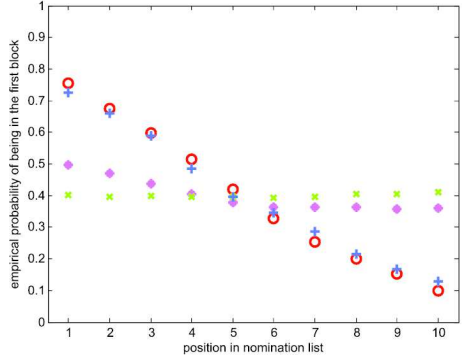}
\\
\footnotesize{(a) Small-scale; $n=10$}
\end{tabular}
\centering
\begin{tabular}{@{}cc@{}}

\includegraphics{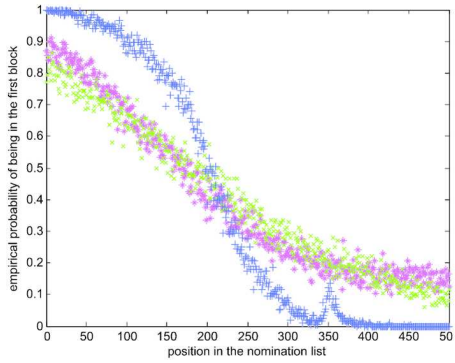}  & \includegraphics{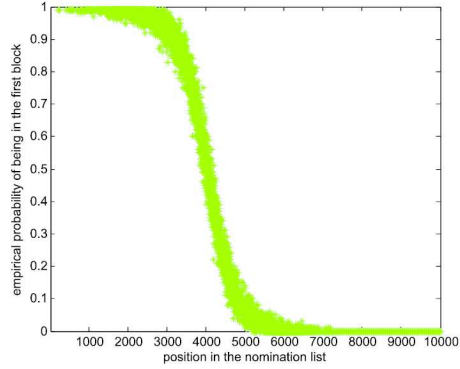}\\
\footnotesize{(b) Medium-scale; $n=500$} & \footnotesize{(c)
Large-scale $n=10\mbox{,}000$}
\end{tabular}
\caption{The canonical vertex nomination scheme is in red, the
likelihood maximization vertex nomination scheme is
in blue, the OTS vertex nomination scheme is in purple,
and the spectral partitioning vertex nomination scheme is in green.
(Canonical scheme not shown in medium- and large-scale figures,
liklihood maximization and OTS schemes
not shown in large-scale figure.)}\label{fig:sims}
\end{figure}

These three experiments were performed as follows.
We independently realized 50,000 graphs from the
associated distribution of the small-scale experiment,
$200$ graphs in the medium-range experiment
and $100$ graphs in the large-scale experiment.
To each observed graph we applied
each of the following: the canonical vertex nomination scheme $\Phi^{C}$,
the likelihood maximization vertex nomination scheme $\Phi^{L}$,
the OTS vertex nomination scheme $\Phi^{O}$ and the
spectral partitioning vertex nomination scheme $\Phi^{S}$.
Then, for each vertex nomination scheme, we recorded the fraction of
the realizations for which the first nominee of the nomination list
was a member of the block of interest, the fraction of
the realizations for which the second nominee
was a member of the block of interest, \ldots, the fraction of
the realizations for which the $n$th nominee
was a member of the block of interest. In Figure~\ref{fig:sims}(a),
(b) and (c) these
empirical probabilities are plotted against nomination list position,
for the three respective experiments and the nomination schemes.\looseness=1

In the small-scale experiment, where $n=10$,
the likelihood maximization nomination scheme performed about as well
as the
(``gold standard'') canonical nomination scheme, and the
spectral partitioning nomination scheme performed very poorly---near
chance. Then, in the medium-scale experiment, where $n=500$,
the canonical nomination scheme was no longer practical to compute,
and the OTS and the spectral partitioning nomination scheme performed
nearly as
well as the likelihood maximization nomination scheme.
For a few thousand vertices it was not practical to implement the
likelihood maximization nomination scheme nor OTS, so in the large-scale
experiment, where $n=10\mbox{,}000$, the only nomination scheme that
could be
implemented
was the spectral partitioning nomination scheme.

The empirical mean average precision for the canonical,
likelihood maximization and spectral partitioning
vertex nomination schemes in the three experiments were as
follows (note that the mean average precision for chance is $0.4$):\vspace*{15pt}

\noindent\tabcolsep=0pt
\begin{tabular*}{\textwidth}{@{\extracolsep{\fill}}lcccc@{}}
\hline
\textbf{Mean average precision} & \mbox{\textbf{Canonical}} & \mbox{\textbf{Likeli-max}}
& \mbox{\textbf{OTS}} & \mbox{\textbf{Spectral}}
\\
\hline
\mbox{Small-scale exper.}, $n=10$, $\vartheta=1$
& 0.6958 & 0.6725 & 0.4763 &0.3993 \\
\mbox{Medium-scale exper.}, $n=500$, $\vartheta=0.3$
& * & 0.9543 & 0.7846 & 0.7330 \\
\mbox{Large-scale exper.}, $n=10\mbox{,}000$, $\vartheta=0.1$
& * & * & * & 0.9901 \\
\hline
\end{tabular*}

The running times in seconds were as follows:\vspace*{15pt}

\noindent\tabcolsep=0pt
\begin{tabular*}{\textwidth}{@{\extracolsep{\fill}}lcccc@{}}
\hline
\textbf{Running time per simulation} & \mbox{\textbf{Canonical}} & \mbox
{\textbf{Likeli-max.}} & \mbox{\textbf{OTS}} & \mbox{\textbf{Spectral}} \\
\hline
\mbox{Small-scale experiment}, $n=10$ & $\approx0.52$ &
$\approx0.03$ & $\approx0.30$ &$\approx0.01$ \\
\mbox{Medium-scale experiment}, $n=500$ & * &
$\approx332$ & $\approx58$ &
$\approx0.17$ \\
\mbox{Large-scale experiment}, $n=10\mbox{,}000$ & *
& * &
* & $\approx106$ \\
\hline
\end{tabular*}\vspace*{9pt}

Indeed, each of the canonical, likelihood maximization
and spectral vertex nomination schemes is superior
(in the sense of effectiveness,
given practical computability limitations)
to the other two at one of the
three scales.
At a small scale you should use the canonical vertex nomination scheme,
at a medium scale you should use the likelihood maximization
vertex nomination scheme, and at a large scale you
should use the spectral partitioning vertex nomination scheme.

\section{Real data examples}
\label{realdata}

While the stochastic block model is often useful for modeling real
data, many times real data does not fit the model particularly well.
In the following real-data experiments we see that
the likelihood maximization vertex nomination scheme
is robust to the lack of idealized conditions hypothesized and other
pathologies inherent in real data. All of the data and code used in
these experiments
can be accessed at \surl{http://www.cis.jhu.edu/\textasciitilde parky/vn/}.

\subsection{Example: The enron graph}
\label{enron}

The Enron Corporation was a highly regarded,
large energy company that went spectacularly bankrupt in the early 2000s
amid systemic internal fraud. Enron has since become a
popular exemplar of corporate fraud and corruption. In the wake of
Enron's collapse, the US Energy Regulatory Commission collected a
corpus of more than $600\mbox{,}000$ emails sent between Enron employees,
and this corpus was made public by the US Department of Justice and
is available online at a number of websites, including
\surl{http://research.cs.queensu.ca/home/skill/siamworkshop.html}.

\begin{figure}
\centering
\begin{tabular}{@{}c@{}}

\includegraphics{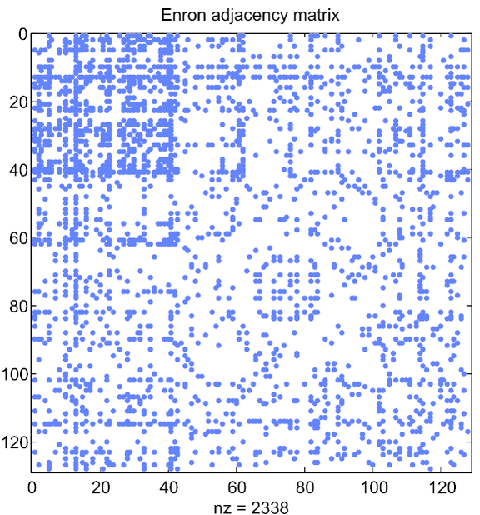}
\\
\footnotesize{(a)}
\end{tabular}
\centering
\begin{tabular}{@{}c@{\quad}c@{}}

\includegraphics{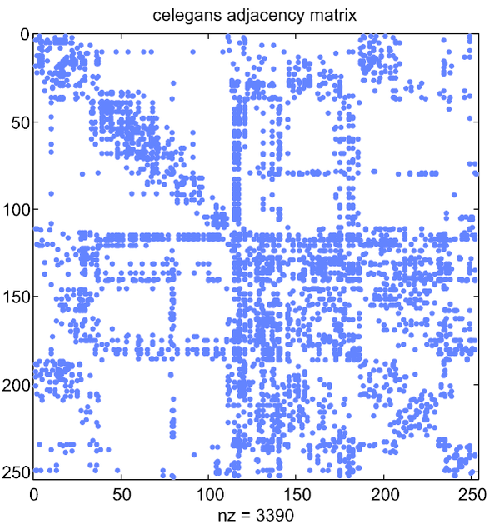}  & \includegraphics{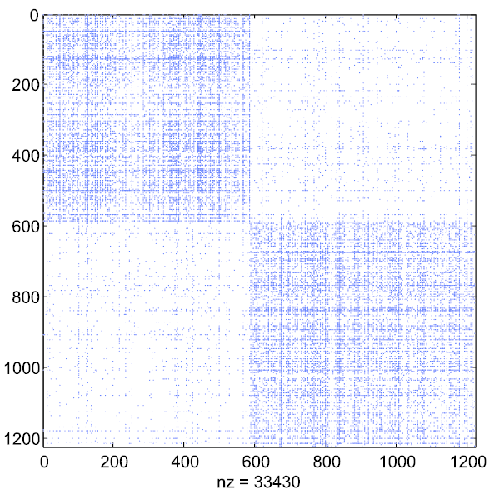}\\
\footnotesize{(b)} & \footnotesize{(c)}
\end{tabular}
\caption{Adjacency matrices for the real-data examples
of Section \protect\ref{realdata}.
\textup{(a)} Vertices partitioned into the 43 upper-echelon employees,
then the 85 lower-echelon employees.
\textup{(b)} Vertices partitioned into the 110 motor neurons, 76
interneurons, then 67 sensory neurons.
\textup{(c)} Vertices partitioned into the $588$ liberal blogs, then
the $636$ conservative blogs.}\label{fig:adjac}
\end{figure}

%
%

In \citet{priebe2005scan}, the authors restrict their attention to a
$189$ week period from the year $1998$ through the year $2002$;
they identify $184$ distinct email addresses in the Enron email
corpus over this time interval, and they identify the pairs of these
email addresses that had email communication with each other.
Our ``Enron Graph'' that we use here is based on the graph in
\citet{priebe2005scan}; our vertex set $W$ consists of the $128$ active
email addresses
for which the employee's job title in Enron was known.
For every pair of such vertices, the pair of vertices are declared
adjacent to each other when there was at least one email sent from
either of the
email addresses to the other. We then divided the vertices into two
blocks: The ``upper-echelon'' set of vertices
$\{ w \in W: b(w)=1 \}$ are the vertices whose job titles were
designated as
CEO, president, vice president, chief manager,
company attorney and chief employee. The ``lower-echelon'' set
of vertices $\{ w \in W: b(w)=2 \}$ are the vertices whose job titles
were designated
as employee, employee administrative,
specialist, analyst, trader, director and manager (besides chief manager,
which we designated upper echelon). We chose to
group the job titles of manager and director with lower-echelon because
a by-eye assessment of the graph indicated that their adjacency
affinity was closer to the rest of the lower-echelon vertices.
Indeed, this graph is certainly not a realization of an actual
two-block stochastic block model, but for the purpose of illustration
we will view it as very roughly having some two-block structure. The adjacency
matrix is pictured in Figure~\ref{fig:adjac}(a).

\begin{figure}

\includegraphics{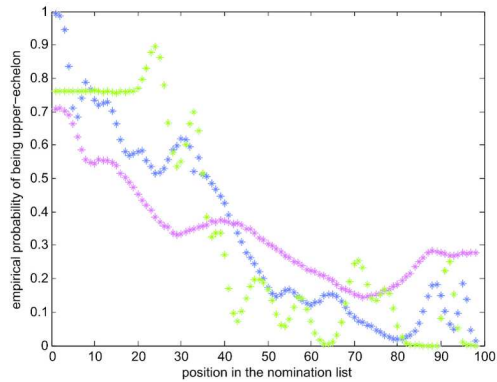}

\caption{Vertex nomination in the Enron Graph. The likelihood maximization,
spectral partitioning and OTS vertex nomination schemes are, respectively,
in blue, green and purple.}\label{fig:the}
\end{figure}

We consider the following experiment. From the $43$ upper-echelon
vertices $\{ w \in W: b(w)=1 \}$, discrete-uniform randomly select
$m_1=10$ to have their block labels known, and the remaining $n_1=33$
have their block labels obscured. From the $85$ lower-echelon
vertices $\{ w \in W: b(w)=2 \}$, independently, discrete-uniform
randomly select
$m_2=20$ to have their block labels known, and the remaining $n_2=65$
have their block labels obscured. Then compute
$\hat{\Lambda}_{1,1}$, $\hat{\Lambda}_{2,2}$
and $\hat{\Lambda}_{1,2}$ as, respectively, the
number of edges in the graph induced by the known
upper-echelon vertices, the number of edges in the
graph induced by the known lower-echelon vertices, and the
number of edges in the bipartite graph induced by the known
upper-echelon and the known lower-echelon vertices, divided, respectively,
by ${n_1 \choose2}$, ${n_2 \choose2}$ and $n_1n_2$.
Then perform likelihood maximization and spectral partitioning vertex nomination
on this graph, using $\hat{\Lambda}$ as a
substitute for $\Lambda$.

We independently repeated this experiment 30,000 times;
Figure~\ref{fig:the} plots the empirical probabilities
of vertex membership in the upper echelon for the respective $98$ positions
in the nomination list, using the likelihood maximization vertex nomination
scheme (in blue), the OTS vertex nomination scheme (in purple)
and the spectral partitioning vertex nomination schemes (in green).
These three vertex nomination schemes had empirical mean average precisions
$0.7779$ (likelihood maximization), $0.7619$ (spectral partitioning)
and $0.5970$ (OTS). For comparison, the mean average precision of chance
is $0.3367$.

Note here that the overall classification success of
spectral partitioning (i.e., the nominating success
averaged over the first 33 positions of the nomination list)
is seen in Figure~\ref{fig:the} as being comparable to
the classification success of likelihood maximization.
Also, here the mean average precision of spectral partitioning
nomination is comparable to that of likelihood maximization nomination.
However, here, very near the top of the nomination list, there is a
visible plateau
in the spectral partitioning nomination success, whereas
maximum-likelihood is nominating very well; indeed, the first few
nominees are nearly always from the block of interest.

\subsection{Example: The caenorhabditis elegans connectome}
\label{celg}

The \textit{Caenor}-\break \textit{habditis elegans} (C.elegans) is a small roundworm whose
connectome\break (neural-wiring) has been completely mapped out;
see \surl{http://www.\\openconnectomeproject.org/\#!celegans/c5tg}.
Our graph here has vertex set $W$ consisting of the
$253$ nonisolated neurons and, for every pair of vertices, the two
vertices are
defined to be adjacent to each other if they are adjoined by a chemical synapse.
Each neuron (i.e., vertex) is exactly one of the following neuron types:
motor neuron, interneuron, or sensory neuron. For each $w \in W$, we define
the block membership $b(w)$ to be $1,2,3$, respectively, according to
whether the neuron
is a motor neuron (there are $110$ of these),
interneuron (there are $76$ of these) or sensory neuron (there are $67$
of these).
The adjacency matrix is pictured in Figure~\ref{fig:adjac}(b).

Consider the following experiment.
Block membership is revealed for $30$ discrete-uniformly
selected motor neurons, $20$ discrete-uniformly
selected interneurons and $10$ discrete-uniformly
selected sensory neurons.
We are interested in forming a nomination list out of the remaining
$193$ ambiguous neurons so that the
beginning of the nomination list has an abundance of (the remaining $80$)
ambiguous motor neurons.

Perhaps the story behind your desire for this nomination
list might be that you wish to study motor neurons,
but have limited resources to biochemically test neuron type for
the ambiguous neurons. The nomination list would be used to order the
ambiguous neurons for the testing, to identify as many motor
neurons as possible from the ambiguous neurons before
your resources are depleted.

\begin{figure}[b]

\includegraphics{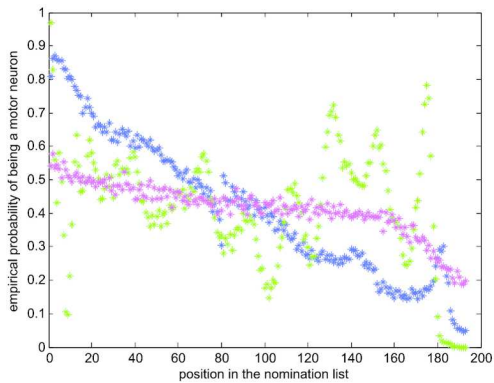}

\caption{Vertex nomination for motor neurons in C. Elegans:
Likelihood maximization is colored blue, OTS is colored purple,
spectral partitioning is colored green.}
\label{fig:VN_on_celegans}
\end{figure}

We repeated this experiment $1000$ times,
each time nominating for motor neurons using
the likelihood maximization, the spectral partitioning vertex nomination
scheme and the OTS vertex nomination scheme.
In each repetition, we estimated $\Lambda$ with
$\hat{\Lambda}$, whose entries reflect the edge densities
in the subgraphs induced by
the various blocks intersecting the seeds.
The empirical mean average precision
for the likelihood maximization, spectral partitioning and
OTS vertex nomination schemes were, respectively,
$0.7272$, $0.5096$ and $0.5041$; the mean average precision of chance
is $0.4145$. Figure~\ref{fig:VN_on_celegans} shows that
empirical probability of being a motor neuron at every
position in the vertex nomination list, for the
likelihood maximization (blue), OTS (purple) and
spectral partitioning (green) vertex nomination schemes.

Note that here spectral partitioning performed very erratically
and (overall) poorly. This might be attributed to a lack of our
idealized three-block structure here; that is to say, this graph does
not appear
to be an instantiation of monolithic
stochastic behavior for vertices within the respective three blocks.
In this case here,
likelihood maximization was seen to be more robust to the lack
of idealized block model setting, and still maintained a steady and
very pronounced slope in Figure~\ref{fig:VN_on_celegans}.

\subsection{Example: A political blog network}
\label{blog}

The political blogosphere data in our next example was
collected in \citet{Adamic:2005:PBU:1134271.1134277} around the time of
the US presidential
election in $2004$. This data set consists of $1224$ weblogs (``blogs''),
each of which web-links to---or is web-linked from---at least one other
of these blogs. These blogs form the vertex set $W$ of our graph. Each
of the blogs was classified by \citet{Adamic:2005:PBU:1134271.1134277}
as being
either liberal or conservative; for each $w \in W$ we define
$b(w)$ to be $1$ or $2$, according to whether $w$ was classified as liberal
or conservative. There are $588$ liberal blogs and $636$ conservative
blogs here. For each pair of vertices/blogs, the pair is
adjacent if at least one of the blogs links to the other.
The adjacency matrix is pictured in Figure~\ref{fig:adjac}(c).

Consider the following experiment. Discrete-uniform randomly
select $80$ liberal and $80$ conservative blogs to have their political
orientation revealed, and create a nomination list for
the remaining $1064$ ambiguous blogs. The story could be that you work for
a political action committee and want to make a report
summarizing liberal blog views on some current event.
You have a limited amount of blog-reading time and only know the
content and political affiliations
of a few of the blogs. Thus, you want to create a
nomination list which will provide the order for
your reading the ambiguous blogs, so that you read many liberal blogs
in your limited time.

\begin{figure}

\includegraphics{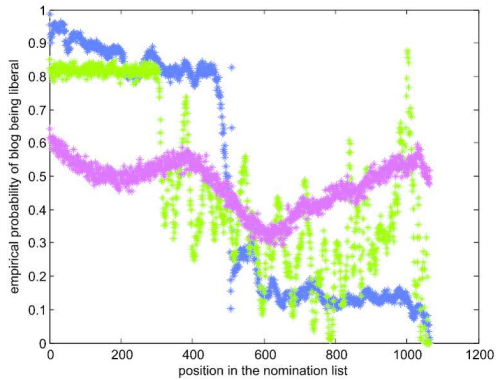}

\caption{Nominating blogs; likelihood maximization vertex nomination
scheme is
colored blue, OTS vertex nomination scheme is colored purple, and
spectral partitioning vertex nomination scheme is colored green.}
\label{fig:cccm}
\end{figure}

We repeated this experiment $1000$ times and
calculated the likelihood maximization,
spectral partitioning and OTS vertex nomination schemes for each
repetition. See the results in Figure~\ref{fig:cccm}.
The mean average precision for the likelihood maximization,
spectral partitioning and OTS vertex nomination schemes were, respectively,
$0.8922$, $0.7856$ and $0.5429$; the mean average precision for chance
nomination was $0.4774$.

\section{Real-data example: Memex and human-trafficking}
\label{memex}

The Defense Advanced Research Projects Agency (DARPA) is an
agency of the United States Department of Defense which, historically,
was responsible for developing computer networking and NLS
(an acronym for ``oN-Line System''), which was the first
hypertext system and an important precursor to the contemporary
graphical user interface (Wikipedia, The Free Encyclopedia, ``DARPA'', accessed
February 15, 2015).

Today's web searches use a centralized, one-size-fits-all approach,
which is very successful for everyday, common use.
DARPA launched the Memex (a contraction of ``Memory Extender'') Program
to create domain-specific index and search, which promises to be a substantially
more powerful search tool, due to its domain specificity. The first
domain that Memex
has addressed is the general domain of human trafficking, which is an
important problem for law enforcement, as well as the military and
national intelligence services. Forums, chats, advertisements,
job postings, hidden services, etc., on the web continue to enable
a growing industry of modern slavery. The index curated by Memex for the
counter-trafficking domain includes a rich set of data with millions of
attributes that, when analyzed with technology, can show linkages between
content that are not easily discoverable by a human analyst.

The graph $G$ that we now consider can be accessed at
\surl{http://www.cis.jhu.\\edu/\textasciitilde parky/vn/} and
is associated with the DARPA Memex and XDATA programs.
It has 31,248 vertices; each vertex corresponds to a web
advertisement. For each pair of vertices, the pair are defined to be
adjacent if the return contact information of the respective advertisements
either share a return phone number or share a return address region
(i.e., city/municipality/metropolitan area). There were 12,387 nodes whose
advertisements had a particular string in the web URL which was ubiquitous
to activities associated with human trafficking; these vertices were
designated ``red.'' The remaining 18,861 vertices were designated ``nonred,''
and it remains unknown if the associated advertisements have any
association whatsoever with human trafficking.

%

The broad goal is, of course, to identify nonred vertices/advertisements
that have association with human trafficking. The direct approach
of forming one large nomination list of the 18,861 nonred vertices is
complicated; among the vertex nomination schemes introduced here, only
the spectral partitioning nomination scheme is practical to directly
compute for a graph this large, and the spectral partitioning
is almost entirely ineffective (the adjusted rand index [\citet
{hubert1985comparing}] between red/nonred
and $k$-means on a two-dimensional embedding was $0.00707$).
Also, keeping in mind the benefits of model averaging,
we decided to perform 10,000 independent replicates
of the following smaller-scale procedure, using likelihood maximization
nomination:

We discrete-uniformly randomly sampled 125 red vertices from among
the 12,387 red vertices, and then discrete-uniformly sampled 50 of
these 125 red vertices to be seeds (their status as red revealed for
what follows) and the other 75 to be ambiguous (their status as red
deliberately and temporarily obscured for what follows).
We then also discrete-uniformly randomly sampled 125 nonred vertices
from among the 18,861 nonred vertices, and then discrete-uniformly
sampled 50 of these 125 nonred vertices to be seeds (their status as
nonred revealed for what follows) and the other 75 to be ambiguous
(their status as nonred deliberately and temporarily obscured for what
follows). We then used the likelihood maximization vertex nomination scheme
to nominate the $150$ ambiguous vertices (among the $250$ selected).

\begin{figure}[b]
\centering
\begin{tabular}{@{}c@{\quad}c@{}}

\includegraphics{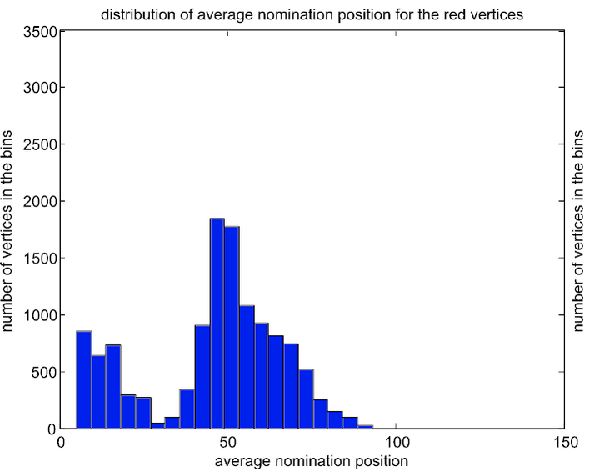}  & \includegraphics{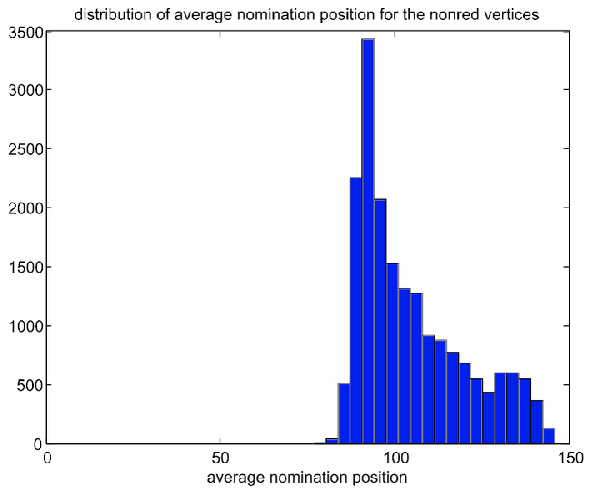}\\
\footnotesize{(a) Red vertices of Memex graph} & \footnotesize{(b)
Nonred vertices of Memex graph}
\end{tabular}
\caption{Histograms of average nomination position
for red vertices and nonred vertices in Memex.}\label{fig:memexfig}
\end{figure}

For each of the 10,000 replications of the procedure described
in the preceding paragraph, we noted the nomination position
(from $1$ to $150$) of each of the ambiguous vertices and, for
each of the 31,248 vertices of the graph, we averaged the vertex's
nomination position over the many times that the vertex was
selected to be ambiguous. In Figure~\ref{fig:memexfig}(a) we plotted a histogram
of the 12,387 red vertices, binned according to average
nomination position, and in Figure~\ref{fig:memexfig}(b) we plotted a histogram
of the 18,861 nonred vertices, binned according to average
nomination position. Note that some of the nonred vertices are
much more likely to appear higher in the nomination lists than
other nonred vertices; the left spike in the histogram of Figure~\ref{fig:memexfig}(b) identifies nonred vertices that should
have a higher priority for scrutiny to ascertain if they are associated
with human trafficking. This outcome is of operational significance.

\section{Discussion}
\label{disc}

In this paper the currently-popular stochastic block\break model setting
enables the principled development of vertex nomination\break schemes.
We introduced several vertex nomination schemes: the canonical,
likelihood maximization, spectral partitioning and OTS vertex nomination
schemes. In Section~\ref{sims} we compared and
contrasted the effectiveness and runtime of these
vertex nomination schemes at small, medium and large scales.
In Proposition~\ref{optim} we proved that
the canonical vertex nomination scheme has maximum
possible mean average precision among all vertex nomination schemes,
and thus it should be used
as long as it is computationally feasible, which is up to a few tens
of vertices. (The runtime visibly grows exponentially in the number of
vertices.)
The likelihood maximization vertex nomination scheme, which
utilizes state-of-the-art graph matching machinery,
should be used next (i.e., when the canonical vertex nomination scheme
can not be used), as long as it is computationally feasible,
which is up to around $1000$ or $1500$ vertices.
Sections \ref{enron}, \ref{celg} and \ref{blog}
then feature illustrations with real data and illustrate
robustness of maximum-likelihood nomination to model pathology
inherent in real data. Section~\ref{memex} highlights an
important contemporary application to stopping human trafficking.

These vertex nomination schemes are simple, yet effective.
The likelihood maximization, spectral partitioning and OTS
vertex nomination schemes are grown from basic block estimation
strategies. Going forward, we expect to see the next generation of
vertex nomination schemes build on similar such adaptations of
block estimation strategies.

\section*{Acknowledgments} The authors thank
Daniel Sussman, Stephen Chestnut and Todd Huffman
for valuable discussions, and
the Editors and anonymous referees for very thoughtful suggestions that
greatly improved the paper.

%

\printaddresses
\end{document}